\documentclass{article}

\usepackage[final]{corl_2019} 

\usepackage{graphics} 
\usepackage{graphicx}
\usepackage{float}
\usepackage{epsfig} 
\usepackage{amsmath} 
\usepackage{amssymb}  
\usepackage{textcomp}
\usepackage{booktabs}
\usepackage{multirow}
\usepackage{hyperref}

\usepackage{color,soul}
\usepackage[nomargin,inline,marginclue,draft]{fixme}

\DeclareMathOperator*{\argmax}{arg\,max}
\DeclareMathOperator*{\argmin}{arg\,min}

\title{Model-based Behavioral Cloning with \\Future Image Similarity Learning}

\author{
  Alan Wu$^1$, AJ Piergiovanni$^1$, Michael S. Ryoo$^{1,2}$\\
  $^1$Indiana University Bloomington\\
  $^2$Stony Brook University\\
  \texttt{\{alanwu, ajpiergi, myroo\}@indiana.edu} \\
}

\begin{document}
\maketitle


\begin{abstract}
We present a visual imitation learning framework that enables learning of robot action policies solely based on expert samples without any robot trials. Robot exploration and on-policy trials in a real-world environment could often be expensive/dangerous. We present a new approach to address this problem by learning a future scene prediction model solely on a collection of expert trajectories consisting of unlabeled example videos and actions, and by enabling generalized action cloning using \emph{future image similarity}. The robot learns to visually predict the consequences of taking an action, and obtains the policy by evaluating how similar the predicted future image is to an expert image. We develop a stochastic action-conditioned convolutional autoencoder, and present how we take advantage of future images for robot learning. We conduct experiments in simulated and real-life environments using a ground mobility robot with and without obstacles, and compare our models to multiple baseline methods.
\end{abstract}



\section{Introduction}

Learning robot control policies directly from visual input is a major challenge in robotics. Recently, progress has been made with reinforcement learning (RL) methods to learn action policies from images using convolutional neural networks (CNNs) \cite{levine2016learning,zhu2016visnav,finn2017deep,sadeghi2017sim2real,peng2018deepmimic}. Taking advantage of abundant training data (e.g., using a simulator or thousands of robot hours), these approaches provide promising results showing that CNNs within the RL framework enable better state representation learning. Model-based RL works further demonstrated that learning a state-transition model from visual data is able to reduce the number of samples needed for RL. This `state-transition' model capturing how state representations change conditioned on actions is also closely related to `future image prediction' (or representation regression) models in computer vision \cite{walker2014patch,vondrick2016anticipating,oh-action_cond,liu2017video,edenton2018svg,babaeizadeh2018svvp}.

Researchers have also studied imitation learning, taking advantage of expert examples instead of explicit reward signals. These include behavioral cloning (BC) and inverse reinforcement learning (IRL). BC is a supervised policy learning problem: the model is given the expert action taken in a state and attempts to replicate the state-action distribution. On the other hand, IRL attempts to infer the underlying reward signal from expert trajectories and then derives action policies using reinforcement learning. This learns more generalized policies than behavioral cloning \cite{finn2016gcl,sutton2017rl}.

Each of the three research directions mentioned (i.e., model-based RL, behavioral cloning, and IRL-based imitation learning) has its own limitation when applied to a real-world robot task. BC is able to obtain a policy solely based on expert trajectories without any expensive/dangerous on-policy trials. However, due to the amount of expert sample data being inevitably limited, encountering unseen states often leads to a covariate shift that can be irrecoverable without querying the expert (i.e., in unseen states, the predicted action can be unreliable or random) \cite{ross2011reduction}. On the other hand, model-based RL and IRL-based imitation learning still require hundreds of real-world, on-policy trials.  Ideally, we desire learning a real-world action policy from a limited number of expert trajectories, without any further trials, that still generalizes well to unseen states and actions. 

In this paper, we propose a new approach to extend conventional behavioral cloning by learning and taking advantage of a future predictive model that also generalizes to unseen states/actions. Our main idea is to enable (1) learning of a realistic future prediction (i.e., state-transition) model solely from a limited number of expert trajectories (i.e., without any robot exploration), and to make (2) better (zero-trial) action policy learning possible by taking advantage of such transition model.

We design a CNN composed of a stochastic convolutional autoencoder for action-conditioned future image prediction. Given an input frame, the model maps it to a lower-dimensional intermediate representation, which abstracts scene information of the current image. The model then concatenates the action vector, and the convolutional decoder produces an image from this representation. Further, our decoder is designed to take advantage of a stochastic prior estimated using a recurrent neural network (i.e., LSTMs), toward better modeling in stochastic environments. 
Our method uses the model to predict the consequences of taking an action to learn a better action policy. We learn the function evaluating how similar the predicted future image is to expert images jointly with the future prediction model itself. Our action policy directly relies on this evaluation function as the critic, thereby selecting the action that leads to future images most similar to the expert images.

The technical contributions of this paper are: (1) Learning of a realistic future image prediction (i.e., state-transition) model from only expert samples. We show it generalizes to non-expert actions and unseen states. (2) Using the learned state-transition model to improve behavioral cloning by taking advantage of 
(visually) predicted consequences of non-expert actions. Our second component only relies on expert images, similar to zero-shot learning works \cite{torabi2018boc,pathak2018zeroshot}. We conduct our experiments using a ground mobility robot in both real-world office and simulator environments containing everyday objects (e.g., desks, chairs, monitors) without any artificial markers or object detectors.

\section{Related Works}

\vspace{-5pt}
\paragraph{Behavioral cloning and IRL}
Given a state, or an observation such as an image, a behavioral cloning (BC) model learns to produce an action matching one provided by an expert \cite{pomerleau1991efficient}. The main challenge of BC is that it is very difficult to obtain supervised training data that covers all possible test conditions the robot may encounter and thus difficult for it to learn a model that generalizes to unseen states. Techniques such as using additional sensors \cite{pomerleau1989alvinn,giusti2016machine,bojarski2016end} or querying the expert when unseen states are encountered \cite{ross2011reduction} have led to better success. Some have taken a variant approach to BC by only accessing expert observations and deriving resulting actions \cite{liu2017ifo,torabi2018boc}, similar to inverse dynamics. Others have shown improvements to BC by injecting noise into the expert policy during data collection, forcing corrective action to disturbances \cite{laskey2017dart}.

Inverse reinforcement learning (IRL) techniques recover a reward signal from the expert trajectories which are state-action pairs \cite{ng2000irl,abbeel2004apprentice,ziebart2008maxentropy,wulfmeier2015dirl}. The learned reward function is used with traditional RL methods, spending thousands of additional trials to learn action policies matching expert behavior. These methods have recently achieved success with high-dimensional data inputs \cite{finn2016gcl,ho2016mfilpo}. Other works take on an adversarial formulation to imitation learning, such as GAIL \cite{gail2016}. Baram et al. \cite{baram2017imitation} presented a model-based generative adversarial imitation learning (MGAIL) approach that uses fewer expert examples than GAIL.

\vspace{-5pt}
\paragraph{Model-based RL and imitation learning}

One major drawback of reinforcement learning is the large number of online trials needed for the agent to learn the policy. Although some have trained their agent in a simulated environment \cite{sadeghi2017sim2real,zhu2016visnav} to minimize the amount of learning needed in a real-life environment, it is very challenging to design realistic simulation models that transfer well to accurate real-world policies. Model-based RL, utilizing a state-transition model, is able to significantly reduce the number of on-policy samples needed for reinforcement learning. Using convolutional action-conditioned models to predict future scenes \cite{oh-action_cond,chiappa-RES} or robot states \cite{peng2018deepmimic} has been employed in simulated environments. Oh et al. \cite{oh-action_cond} and Chiappa et al. \cite{chiappa-RES} learn transition models using recurrent networks to predict scenes for multiple steps into the future conditioned on a sequence of actions. Peng et al. \cite{peng2018deepmimic} learn an inverse model for multi-jointed humanoid robots to perform complex tasks such as flipping and spinning.

There are also approaches utilizing state-transition model learning (or its inverse model learning) for imitation learning without online trials.
In particular, recent works \cite{torabi2018boc,pathak2018zeroshot} attempted learning of agent policies without expert action labels. The setting was that although state-action pairs from robot exploration trajectories (random or guided) could be used for the state-transition model learning, only the use of expert states (images in our case) were allowed for policy learning. This is similar to our learning process in the sense that no action labels are used in the policy learning stage (i.e., critic learning).

In this paper, we focus on real-world robot action policy learning only using expert trajectories. We combine the concept of stochastic state representation (e.g., \cite{babaeizadeh2018svvp,edenton2018svg}) with the action-conditioned future prediction to generate more realistic future images, and propose the approach to make behavioral cloning learn to benefit from such model in a fully differentiable manner.

\section{Problem Formulation}
We focus on the problem of `zero-trial' imitation learning, which essentially is the setting of behavioral cloning. By `zero-trial' we mean that we learn an action policy of the robot solely based on expert trajectories (i.e., human samples) without any on-policy robot trials. That is, the robot is prohibited from interacting with the environment to obtain its own samples to check/update the policy.
The learned policy is directly applied to the robot in the real-world environment.

We formulate the problem of robot policy learning as the learning of convolutional neural network (CNN) parameters. Our agent (i.e., a robot) receives an image input $I_t$ at every time step $t$. Given $I_t$, our goal is to learn a policy, $\pi$, which given the current state, selects the optimal action to take. That is, the policy is modeled as a neural network, and takes the current image as input and produces an action $a = \pi(I_t)$. In the case of robot imitation learning, $\pi$ has to be learned from expert trajectories (i.e., samples), often obtained by humans controlling the robot. The action $a_t$ could be the direct motor control commands. In this work, we consider continuous state and action spaces where the states are direct images from the robot camera.

\section{Model-based Behavioral Cloning}

We present a zero-trial imitation learning method using a \emph{future image prediction model}. Our future image prediction model is a CNN that predicts a realistic future image given its current image and an action. Our method first learns how the image (from the robot camera) visually changes when the robot takes an action (Section \ref{subsec:future}), and also learns to take advantage of such model for its action policy by evaluating the predicted images (Section \ref{subsec:future-similarity}). 

\subsection{Action-Conditioned Future Prediction CNN}
\label{subsec:future}
\begin{figure}[H]
  \centering
    \includegraphics[width=0.73\textwidth]{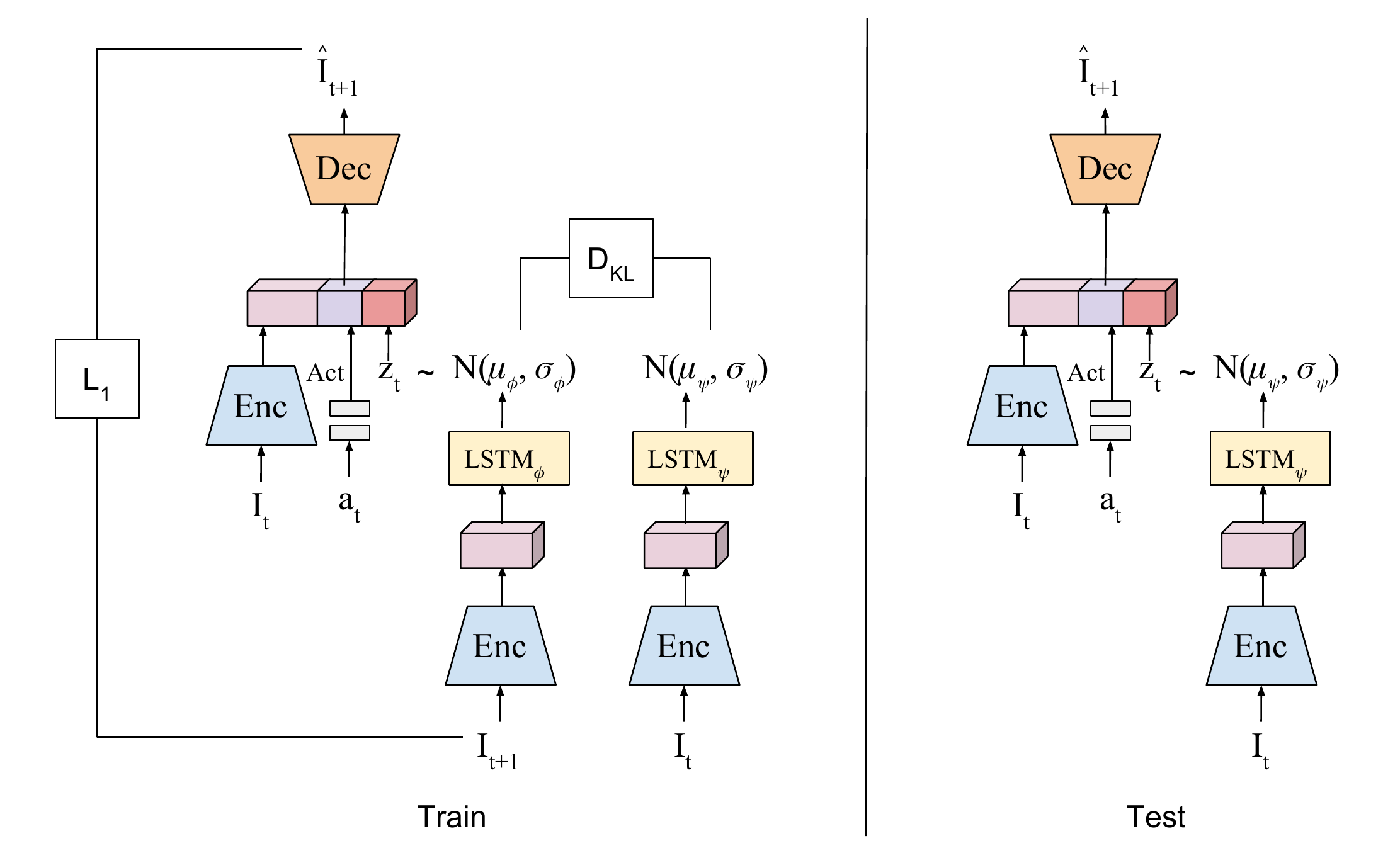}
      \caption{Illustration of the stochastic image predictor model. This model takes as input the image and action, but also learns to generate a prior, $z_t$, which varies based on the input sequence. This is further concatenated with the representation before future image prediction. The use of the prior allows for better modeling in stochastic environments and generates clearer images.}
      \label{fig:model-svg}
\end{figure}

Our CNN architecture is designed to predict future images based on the current image and an action. This can be viewed as a state-transition model used in model-based reinforcement learning, following a Markov decision process (MDP). Further, we learn a prior that captures the stochastic nature of real-world environments motivated by \cite{edenton2018svg}. Our proposed model is shown in Fig.~\ref{fig:model-svg}: the action-conditioned stochastic autoencoder CNN.

The model is composed of two main CNN components: the encoder $Enc$ and the decoder $Dec$. The encoder learns to map the current image to a latent representation, $z_I = Enc(I_t)$. We designed our autoencoder to explicitly preserve the spatial information in its intermediate representations, making $Enc$ and $Dec$ to be fully convolutional, which allows our intermediate scene representation to preserve the spatial shape.

We use another neural network component, $Act$, which learns a representation of the action, $z_a = Act(a_t)$. We also design our action representation CNN $Act$ to have several fully connected layers followed by a reshaping (i.e., spatial de-pooling) layer. This makes the output of $Act$ have the spatial dimensionality as the image representation, i.e., $Enc(I_t)$, which allows the two representations to be concatenated together. Further, this allows for having a different action representation at each spatial location.

In addition, the latent variable $z_t$ is sampled from a learned prior distribution $p_\psi$ also implemented as a neural network, which outputs the parameters of a conditional Gaussian distribution. The latent variable $z_t$ captures the stochastic information in videos that is lacking in deterministic models. A separate inference model $q_\phi$ is used only during training and removed during test. $q_\phi$ is forced to be close to $p_\psi$ using a KL-divergence term, which helps prevent $z_t$ from directly copying $I_t$. As the prior learns stochastic characteristics of video across time from a sequence of images, convolutional LSTMs $LSTM_{\phi}$ and $LSTM_{\psi}$ are used to capture the temporal elements. We use multiple frames as input during training that benefits the learning of the prior, but we use only a single image during robot inference. 

\begin{figure}
  \centering
    \includegraphics[width=0.48\textwidth]{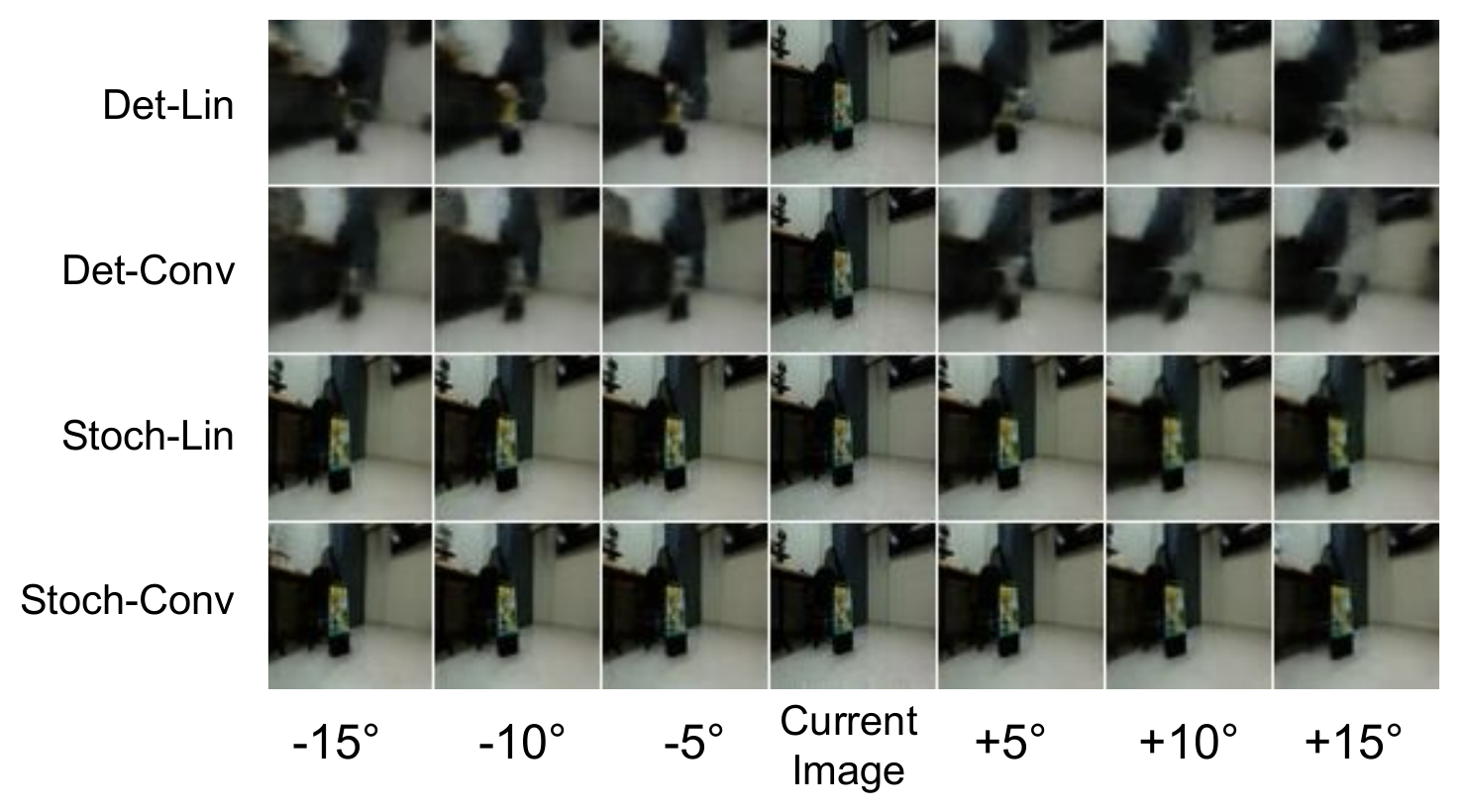}
    \includegraphics[width=0.48\textwidth]{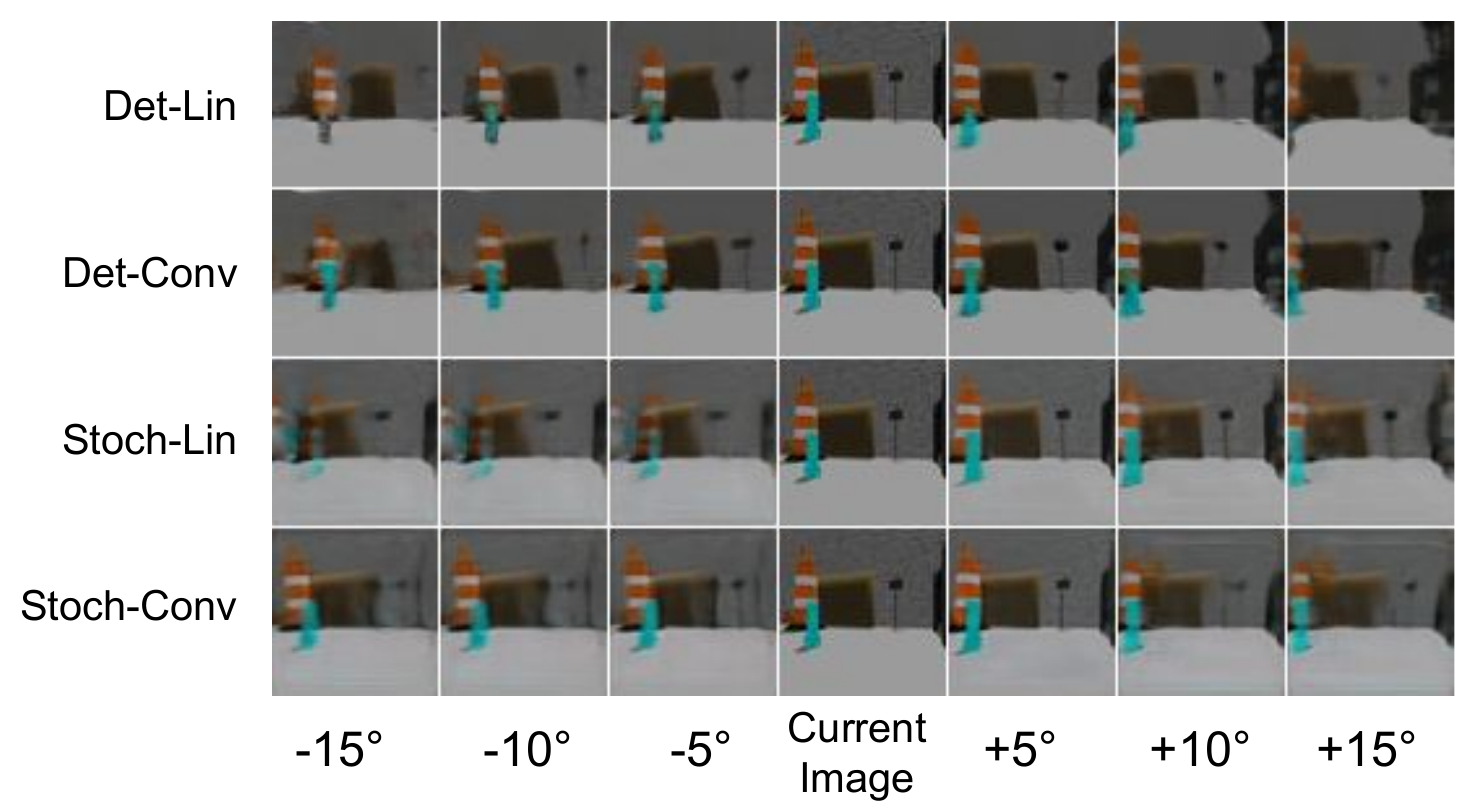}
      \caption{Predicted future images in the real-life lab (left) and simulation (right) environments taking different actions. Top two rows of each environment: deterministic model with linear and convolutional state representation, respectively. Bottom two rows: stochastic model with linear and convolutional state representation, respectively. Center image of each row is current image with each adjacent image to the left turning -5\textdegree \enspace and to the right turning +5\textdegree.}
      \label{fig:future-imgs-arc}
\end{figure}

The representations of the current image, action, and learned prior  are concatenated together $[z_I, z_a, z_t]$ and used as input to the decoder which then reconstructs the future image:
\begin{equation}
       \hat{I}_{t+1} = f(I_t, a_t, z_t) = Dec([Enc(I_t), Act(a_t), z_t])
\end{equation}
As mentioned above, this can also be viewed as learning a state-transition model from robot videos. Fig.~\ref{fig:future-imgs-arc} provides an example of the predicted images in real-life and simulation environments given different actions, visually showing we are able to generate realistic, unseen future images.

The model has two loss terms. It is trained to minimize $L_1$ error between the predicted image and the ground truth future image \cite{isola2016, zhao2015}. It is also simultaneously trained to minimize the KL-divergence between the prior distribution $p_{\psi}$ and an inferred distribution $q_{\phi}$. $\beta$ is a hyperparameter set to 0.0001 as was done in \cite{edenton2018svg}. We used an input sequence of T=5 frames during training and a single frame during test. Multiple frame inputs during test could improve learning and thus a worthwhile pursuit in the future.
\begin{equation}
  \mathcal{L_F}(I_{1:T+1}) = \sum\limits_{t=1}^T\Big(|\hat{I}_{t+1}-I_{t+1}| ~~+~~ \beta D_{KL}[q_{\phi}(z_t|I_{1:t+1})~||~p_{\psi}(z_t|I_{1:t})]\Big)
\end{equation}
Fig.~\ref{fig:model-svg} shows our full model as well as its losses for the training. We emphasize that during training, we only have expert future images as a result of expert actions, greatly limiting the training data. However, as shown in Fig.~\ref{fig:future-imgs-arc}, our learned model is able to generalize to unseen, non-expert actions by generating realistic and accurate images.

\subsection{Action Learning Using Future Image Similarity}
\label{subsec:future-similarity}

To use our learned future prediction model for robot imitation learning, we train a `critic' function to evaluate how similar a generated state-action pair is to an expert state-action pair. The idea is to train a CNN that can distinguish state-action pairs that look like expert samples from non-expert samples, similar to the learning of the `discriminator' in a generative adversarial imitation learning (GAIL) \cite{gail2016}. Learning such a CNN will directly allow for the selection of the optimal action given a state (e.g., image) by enumerating through several action candidates. 

More specifically, we use the above future image prediction to generate images for many different candidate actions. We then learn a critic CNN that evaluates how similar the predicted future image would be to the ground truth, expert future image. We train a CNN to model this function, $\hat{V}$, which performs this evaluation.

To train this CNN, we use pixel-wise $L_1$ difference between the predicted future image and the actual future image (i.e., $|I_{t+1} - \hat{I}_{t+1}|$) as our similarity measure. We train the CNN $\hat{V}(\hat{I}_{t+1}; w)$ governed by the parameters $w$: 
\begin{equation}
\begin{split}
       w^* &= \argmin_{w} \Arrowvert \hat{V}(\hat{I}_{t+1}; w) - |I_{t+1} - \hat{I}_{t+1}| \Arrowvert
\end{split}
\end{equation}
Once learned, given a current robot image $I_t$ and an action candidate $a_t$, we are able to evaluate how good the action candidate is solely based on the predicted future image $\hat{I}_{t+1}$ by computing
\begin{equation}
    \hat{V}(\hat{I}_{t+1} = Dec([Enc(I_t), Act(a_t), z_t]); w^*)
\end{equation} 
The motivation behind this approach is that actions taken by the robot in one state should result in a future image that is similar to the next expert image, if the imitation is being done correctly. $\hat{V}$ reflects this imitation constraint while only taking advantage of image frames. Note that this is possible because we have the explicit future image predictor which allows us to compare two different images (i.e., real expert images vs. model predicted images).

\begin{figure}
  \centering
    \includegraphics[width=0.48\textwidth]{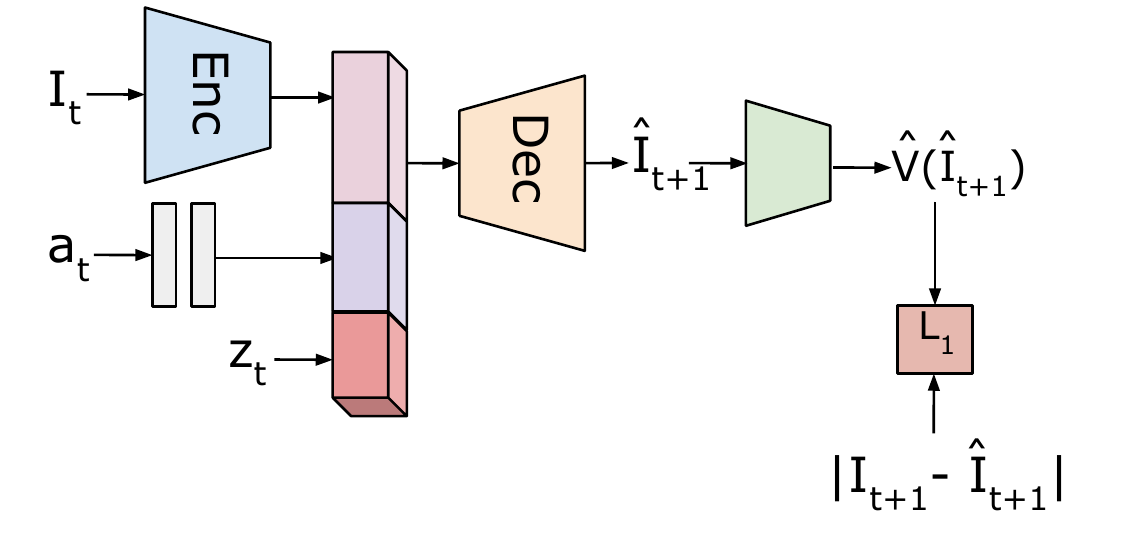}
      \caption{Training of the critic function $\hat{V}$ with the future prediction model for future image similarity. We generate many future images from the current image and various actions, and train the critic to match the $L_1$ difference between the predicted image and the ground truth image.}
      \label{fig:train-image-similarity}
\end{figure}

The training procedure of this method is illustrated in Fig.~\ref{fig:train-image-similarity}. We use our pre-trained future prediction model to obtain the predicted future image from each state-action pair $(I_t,a_t)$ in our training set of expert trajectories. We can then sample many random candidate actions, which contain many non-expert actions.

The main advantage of this approach is the ability to `imagine' unseen future images resulting from non-expert actions taken in expert states. This allows the model to benefit from more training data without further interaction with the environment. Once learned, the optimal action at each state $I_t$ is selected by taking the max value from the randomly sampled candidate actions: $a_t = \argmax_a \hat{V}(f(I_t, a, z_t))$ where
$f(I_t, a, z_t) =  Dec([Enc(I_t), Act(a), z_t])$.

\section{Experiments}
We perform a series of experiments with our method in two different environments, a real-life lab setting and a simulation setting in Gazebo, each with a different task. In the lab, the robot performed an approach task where the robot must move towards a target object. In the simulation environment, the robot must approach a target while avoiding an obstacle placed in between. The latter is a more challenging task as the view of the target from the robot's perspective is partially obstructed by the obstacle and the robot must learn to avoid the obstacle without explicitly detecting it.

\vspace{-5pt}
\paragraph{Datasets}
We collected two datasets of ground mobility robot trajectories: one in a real-world room with everyday objects (e.g., desks, chairs), without any markers, and another in a simulated environment with different obstacles (one at a time), target object, and similar everyday objects. For the lab dataset, on average, we collected 35 100-frame trajectories for each target object, asking a human expert to control the robot to move toward a target object selected from a total of 7 different objects. The initial robot placement and the target location were varied leading to varied background scenery. We split our dataset into 32 training trajectories and 3 test trajectories per target, a total of 224 training trajectories and 21 test trajectories. For the simulation dataset, on average, we collected 100 400-frame trajectories for each obstacle. We split our dataset into 91 training trajectories and 9 test trajectories per obstacle, a total of 182 training trajectories and 18 test trajectories. The objective of our expert trajectory collection was to train our robot. No additional labels (e.g., object annotations) were provided. We perform both offline evaluation and online experiments.

\vspace{-5pt}
\paragraph{Baselines and ablations}
To evaluate our design decisions, we compare our full model against other models that replace or remove some of the components. Specifically, we compared using linear state representations against convolutional state representations (i.e., linear vs. convolutional). Secondly, we examined the impact of the learned prior on our action policy. Without the learned prior, we simply have a deterministic image predictor model that is trained without the latent variable $z_t$ (i.e., deterministic vs. stochastic). We can see in Fig.~\ref{fig:future-imgs-arc} that the predicted images from the deterministic models are blurrier, but wanted to determine if the blurriness had a significant impact on action policy.

In addition to different versions of our approach mentioned above, we implemented a couple of baselines: behavioral cloning and forward consistency \cite{pathak2018zeroshot}. Similar to ours, these models do not perform any learning from interacting with the environment, as opposed to RL models. We provide more detailed descriptions of these baselines in the Appendix. When training/testing \cite{pathak2018zeroshot} based on its publicly available code, we provided an additional input (i.e., the goal image of each task) as required by their model.

\begin{figure}
    \includegraphics[width=0.49\textwidth]{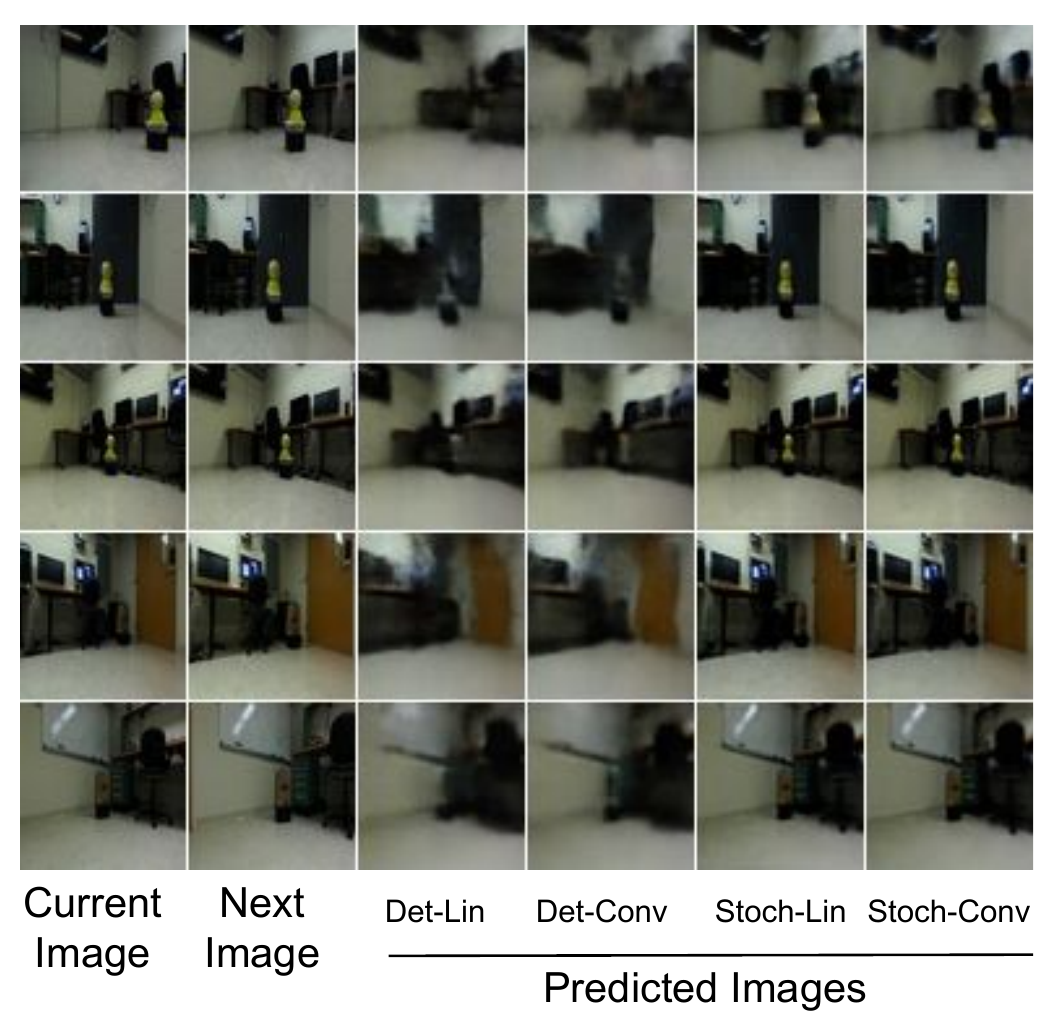}
    \includegraphics[width=0.49\textwidth]{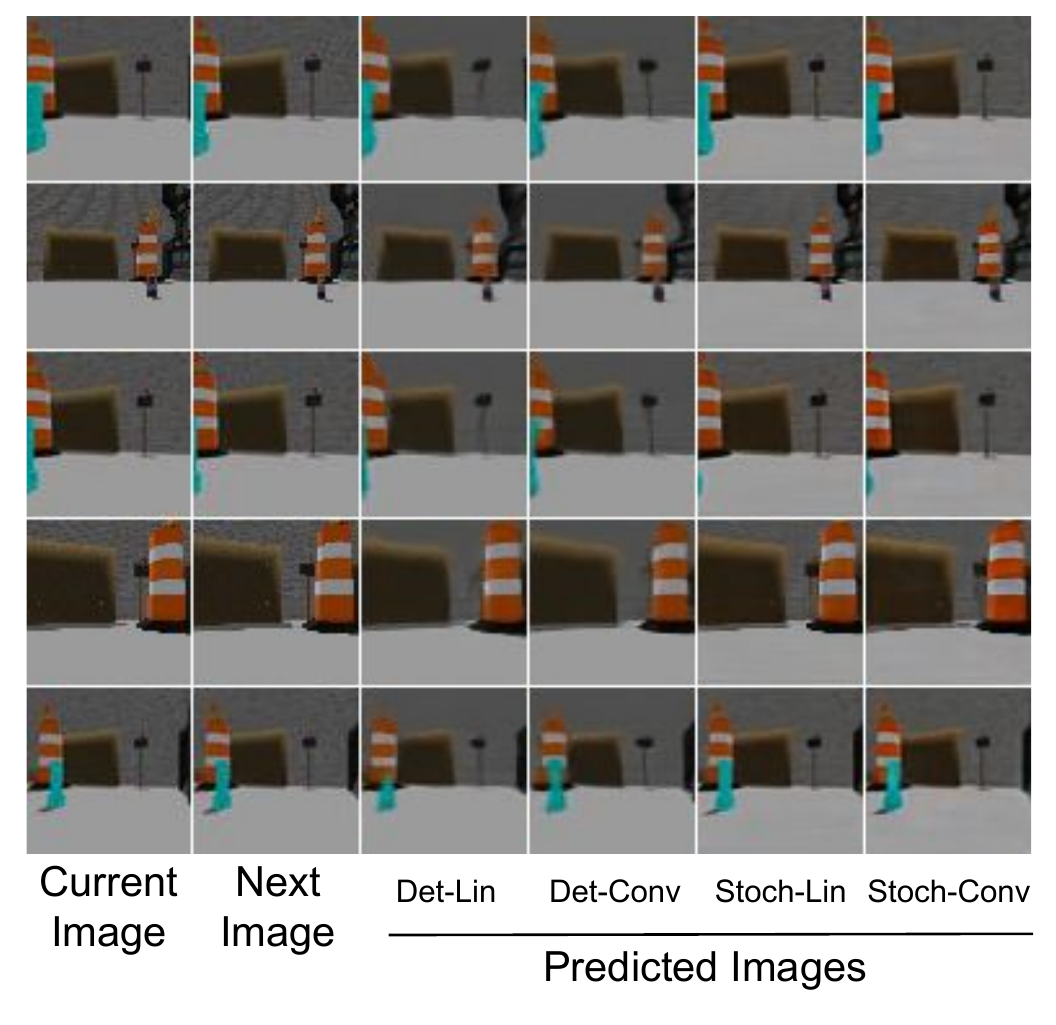}
    \caption{Sample predicted images from the real/simulation dataset (left/right). From left to right: current image; true next image; deterministic linear; deterministic convolutional; stochastic linear; stochastic convolutional. (Left) High level description of action taken for each row starting from the top: turn right; move forward; move forward slightly; move forward and turn left; move forward and turn left.
    (Right) High level description of action taken for each row starting from the top: move forward and turn right; turn right slightly; turn right; move forward slightly; turn left slightly.}
      \label{fig:future_img_comparisons}
\end{figure}

\vspace{-5pt}
\paragraph{Future image model evaluation}
Because our method depends on the quality of the predicted images, we compare the quality of the image predictors using structural similarity (SSIM) \cite{Wang04imagequality,edenton2018svg,babaeizadeh2018svvp}. 
Table~\ref{tab:ssim} 
compares image quality of 2000 generated images of the action-conditioned deterministic and stochastic models, showing both linear and convolutional state representations. Fig.~\ref{fig:future_img_comparisons} allows for visual comparison of the different models, demonstrating that using the learned stochastic prior leads to clearer images. 
We note that the real-world lab dataset benefits more from the convolutional state representation since it has richer scene appearance and more inherent noise that needs to be abstracted into the state representation. On the other hand, the simulation data is much cleaner, sufficient to abstract with lower-dimensional representations. Our stochastic models perform superior to the deterministic models in both datasets.

\begin{table}
\caption{Assessment of image predictor models using SSIM with our real-world lab and simulation datasets. Higher SSIM scores indicate more accurate predictions.}
\label{tab:ssim}
\centering
\begin{tabular}{lcc}
\toprule
Model & Lab Dataset & Sim Dataset \\
\midrule
Det Linear State    & $0.7029\pm0.0086$ & $0.8856\pm0.0036$\\
Det Conv State      & $0.7211\pm0.0082$ & $0.8983\pm0.0037$\\
Stoch Linear State  & $0.7255\pm0.0145$ & $0.9251\pm0.0083$\\
Stoch Conv State    & $0.7436\pm0.0153$ & $0.9220\pm0.0091$\\
\bottomrule
\end{tabular}
\end{table}

\vspace{-5pt}
\paragraph{Robot evaluations}
We conducted a set of real-time experiments with a ground mobility robot in complex environments -- real-world lab and simulation -- to illustrate the implementation of the approaches. For each model in the lab environment, we ran 18 trials for each of three different target objects. For the simulation environment, we ran 27 trials for two different obstacle objects. For the action-conditioned models, each model was given 60 action candidates to choose from, distributed from -30\textdegree{} to +30\textdegree{} relative to the current angular pose of the robot. For each trial, we altered the target object location and the robot starting pose, with the constraint that the target object is within the initial field of view of the robot. We allowed the robot to take up to 30 steps to reach within 0.5 meters of the target. We considered a trial successful if the robot captured an image of the target within the 0.5m. If the robot did not reach the target within 30 steps or if the robot went out of bounds, such as hitting a wall or running into a table, we counted the trial as a failure.  

In Table~\ref{tab:robot-lab} and Table~\ref{tab:robot-gaz}, we show the results of the models we tested in the real-world lab and simulation environments, respectively. Our future similarity models performed superior to standard behavioral cloning and forward consistency \cite{pathak2018zeroshot}, demonstrating the advantage of our action-conditioned future prediction for imitation learning. Our lab experiments indicate that retaining spatial information through convolutional state representation leads to better performance. The component that makes the greatest contribution for improved performance is the learned stochastic prior. Since we found that the main contributor of improvement came from learning the stochastic prior, for our obstacle avoidance task in the simulation environment, we compare only future similarity models using the linear state representation as shown in Table~\ref{tab:robot-gaz}.

\begin{figure*}
    \includegraphics[width=\textwidth]{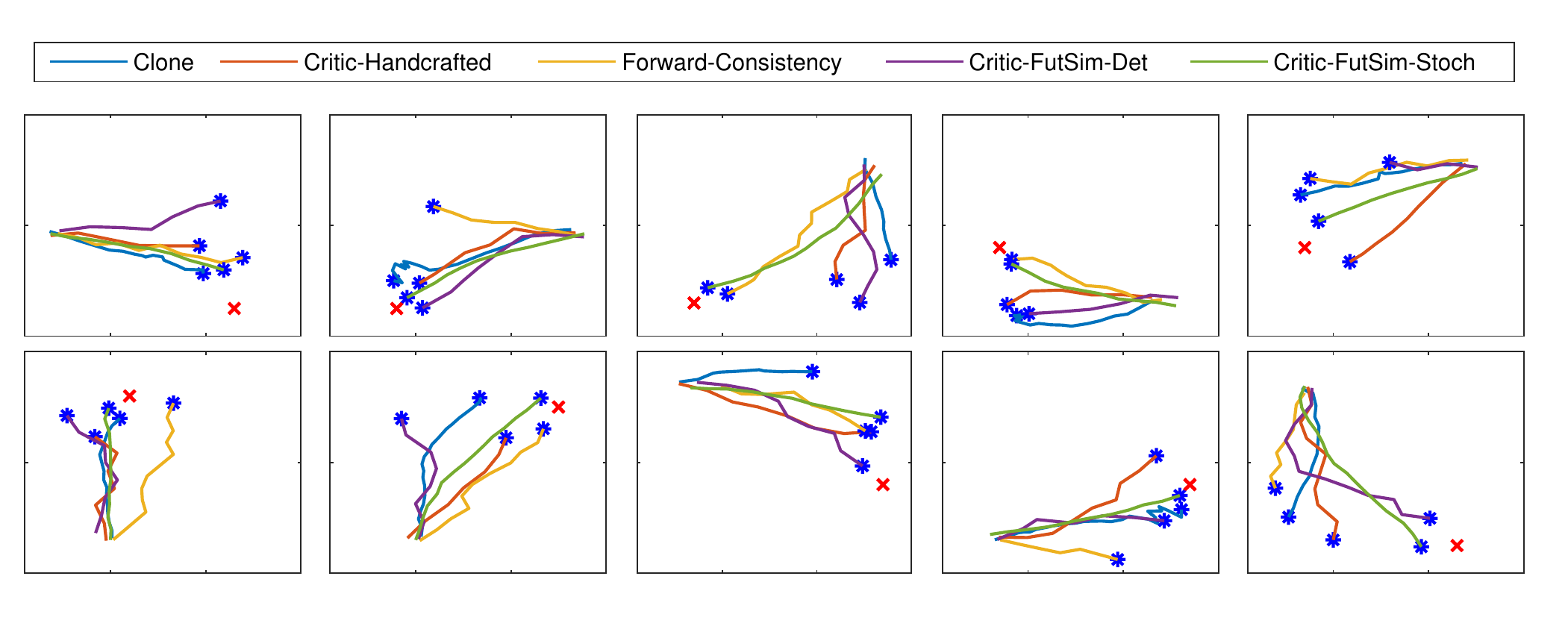}
     \caption{Example trajectories from the real-time robot experiments. The red `$X$' marks the location of the target object and the blue `$*$' marks the end of each robot trajectory. Note that this is a challenging setting, since (1) we only provide a limited number of training examples, (2) it was done in a real-world environment with diverse objects, and (3) we did not provide any annotation of the target object. `Critic-Handcrafted' is a baseline approach trained by providing ground truth distance between the target object and the robot, which we describe more in the Appendix.}
     \label{fig:robot-traj-example}
\end{figure*}

\begin{table}
\scalebox{0.96}{
\begin{minipage}[b]{0.47\linewidth}
\caption{Real-world robot target approach experiment results reporting the task success rate.}
\label{tab:robot-lab}
\setlength\tabcolsep{2pt}
\begin{tabular}{lcccc}
\toprule
Model & Targ1 & Targ2 & Targ3 & Mean \\
\midrule
Behavioral Cloning  & 50\%  & 33\%  & 28\% & 37\% \\
\midrule
Forward Consistency  & 50\%  & 39\%  & 28\%  & 39\% \\
\midrule
Det. Linear State    & 67\%  & 56\%  & 56\% & 59\% \\
Det. Conv State      & 78\%  & 78\%  & 67\% & 74\% \\
Stoch. Linear State  & 94\%  & 89\%  & 89\% & 91\% \\
Stoch. Conv State    & 100\% & 89\%  & 94\% & 94\% \\ 
\bottomrule
\end{tabular}
\end{minipage}
\hspace{0.9cm}
\begin{minipage}[b]{0.49\linewidth}
\caption{Simulated robot obstacle avoidance experiment results reporting the task success rate.}
\label{tab:robot-gaz}
\setlength\tabcolsep{3pt}
\begin{tabular}{lccc}
\toprule
Model & Obst1 & Obst2 & Mean \\
\midrule
Behavioral Cloning   & 41\%  & 48\%  & 44\% \\
\midrule
Forward Consistency   & 37\%  & 41\%  & 39\% \\
\midrule
Deterministic State  & 44\%  & 37\%  & 41\% \\
Stochastic State     & 67\%  & 59\%  & 63\% \\
\bottomrule
\end{tabular}
\end{minipage}}
\end{table}

We also compared the trajectories of the various methods to the expert for 10 different robot-target placements in the lab environment. To measure the similarity between the expert and each of the methods, we use dynamic time warping (DTW) also used in \cite{vakanski-dtw} as a similarity score to assess how close the learned model performed the approach task compared to the expert. DTW is useful for measuring time series of different lengths such as our trajectories by finding the distance of a warp path $W$. If we compare two trajectories ($X_{1:M}$ and $Y_{1:N}$), their similarity using DTW is determined by finding the minimum-distance warp path: $Dist(W) = \sum_{k=1}^K Dist(w_{ki}, w_{kj})$, where $i$ is a point on path $X$, $j$ is a point on path $Y$, $k$ is a point on $W$, and $K = M*N$. $Dist(w_{ki}, w_{kj})$ is the distance between the two data point indices (one from $X$ and one from $Y$) in the $k^{th}$ element of the warp path \cite{Salvador2004FastDTWT}. Table~\ref{tab:similarity} shows that our method achieved the best similarity score.

\begin{table}
\caption{Similarity scores of various models, comparing their generated trajectories to the (unseen) expert trajectory. The lower the better.}
\label{tab:similarity}
\centering
\begin{tabular}{lcc}
\toprule
Model & DTW Score\\
\midrule
Behavioral Cloning   & 28.26\\
Forward Consistency  & 18.36\\
FutureSim-Det        & 11.22\\
FutureSim-Stoch      & 5.98\\
\bottomrule
\end{tabular}
\end{table}

Our method is also robust to distractor objects. We conducted an experiment where we placed a distractor object in the scene about equidistant to the robot as the target object. We varied the target object, the distractor object, and their locations with respect to the robot. Even without any labeling of the target or distractor objects, our robot is able to ignore the distractor and navigate to the target based on the learned models. A similarity score of 4.94 was achieved using the stochastic model, reflecting consistently small deviation from the expert. In the Appendix, we show examples of the target in the lab environment with and without distractors as well as example trajectories.

\section{Conclusion}

We presented a model-based visual imitation learning framework that uses the same amount of training data as behavioral cloning. The goal was to make a real-time robot learn to execute a series of actions in real-world and simulation environments (e.g., an office) solely based on its visual expert samples without benefiting from any further on-policy trials. The learning was done without providing any explicit labels showing what the target object looks like, relying only on a small set of example videos. We introduced an approach to learn a stochastic action-conditioned convolutional future image prediction model (i.e., a visual state-transition model). We experimentally confirmed that our model benefits from a learned prior to capture stochastic components of real-world videos and, in turn, generates clearer images. We further confirmed that our future image similarity model that learns to evaluate the predicted future of each action successfully allows imitation of the expert trajectories, better generalizing to unseen states and actions. Our model achieved 2.5x higher success rate in the real-world robot approach task than the previous methods.

\subsubsection*{Acknowledgments} This work was supported by the Institute of Information \& Communications Technology Planning \& Evaluation (IITP) grant funded by the Korean government (MSIT) (No. 2018-0-00205, Development of Core Technology of Robot Task-Intelligence for Improvement of Labor Condition).

\bibliography{ref}  

\appendix
\section{Appendix}

\subsection{Dataset Details}

\begin{figure}[H]
    \centering
    \includegraphics[width=0.2\textwidth]{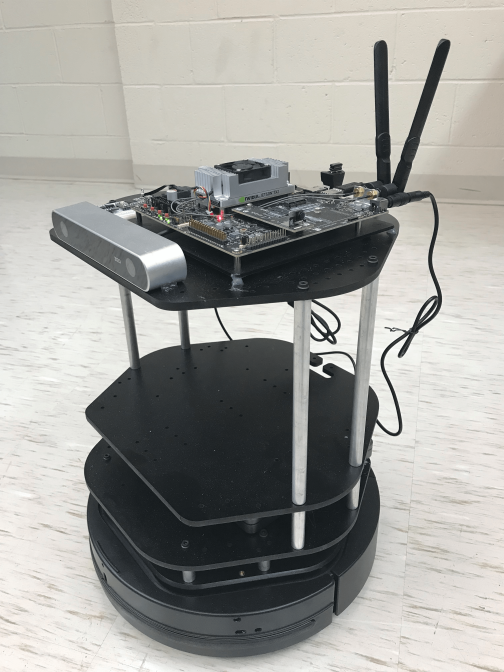}
    \hspace{1mm}
    \includegraphics[width=0.4\textwidth]{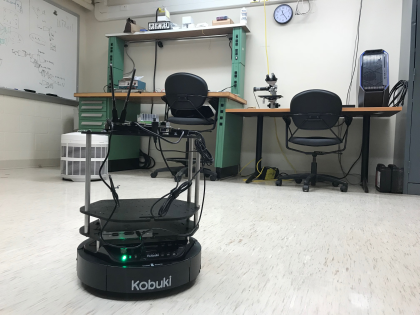}
    \caption{Images of our ground mobility robot (Turtlebot 2) and its environment.}
    \label{fig:turtlebot}
\end{figure}

\begin{figure}[H]
  \centering
    \includegraphics[width=0.4\textwidth]{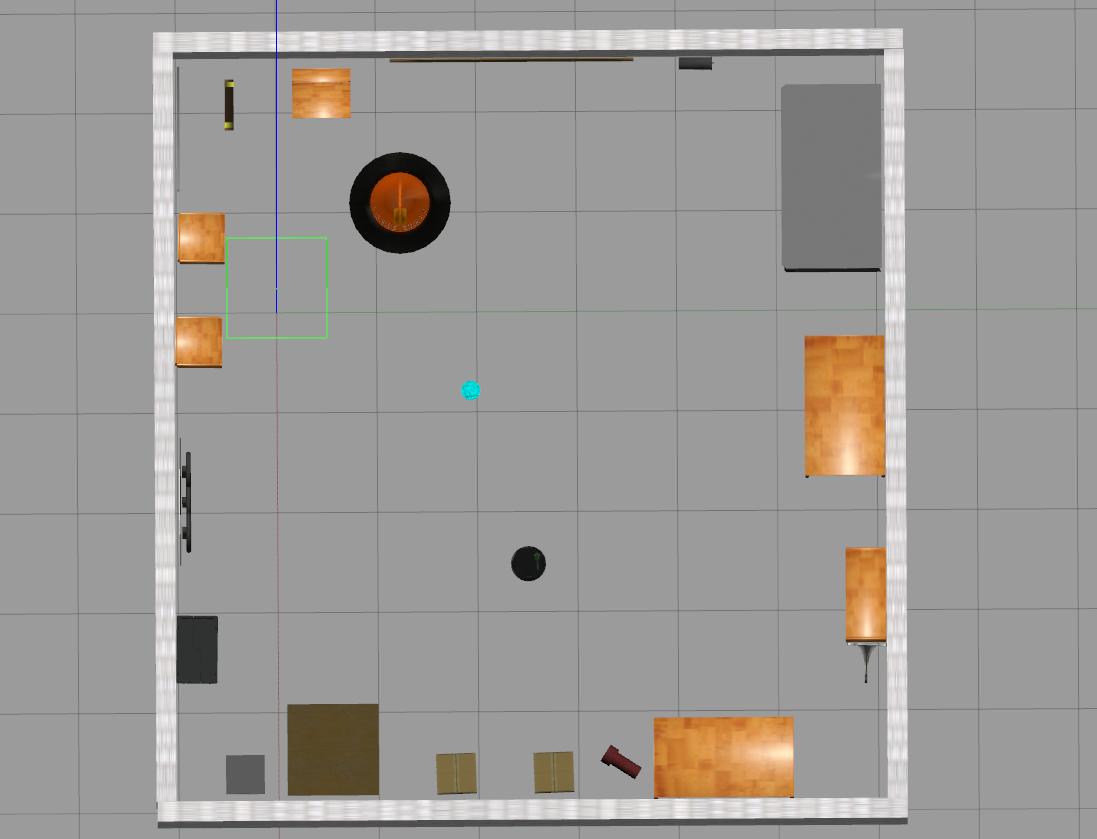}
    \includegraphics[width=0.3\textwidth]{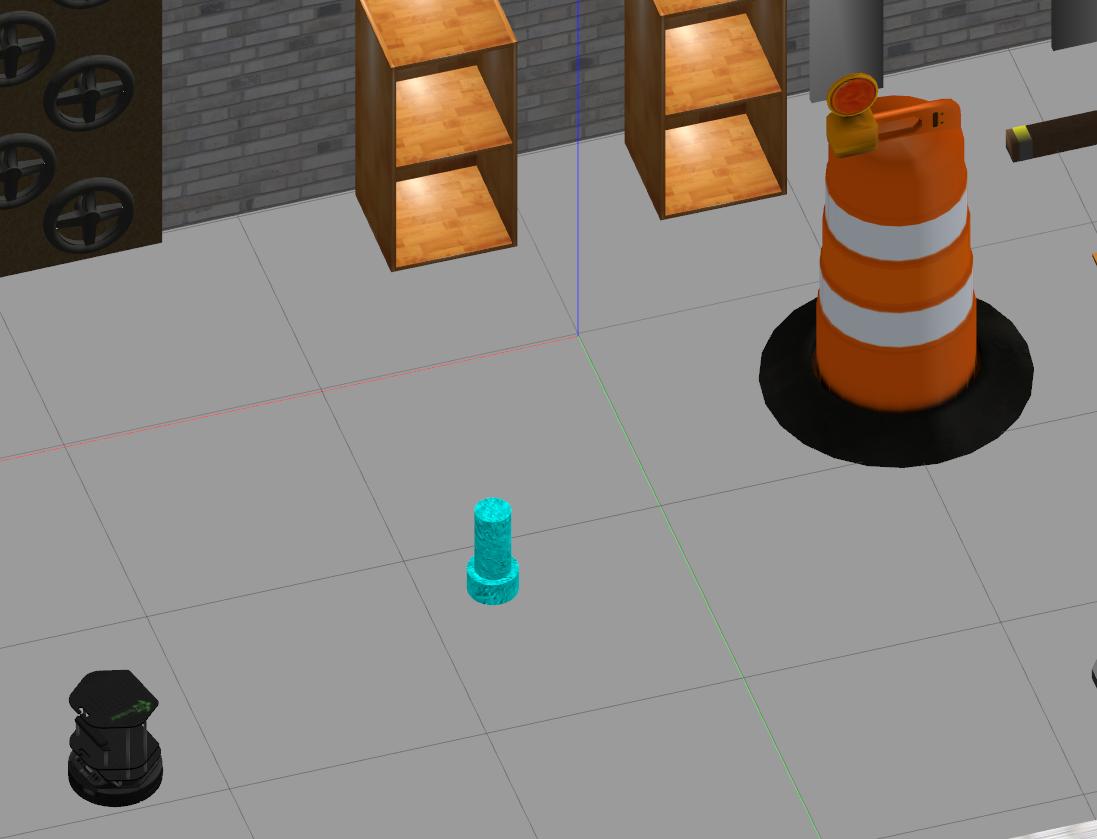}
      \caption{Our simulation environment in Gazebo: birdseye view (L) and side view (R). An obstacle (blue cylinder) is placed between the robot and the target (construction barrel).}
      \label{fig:gazebo}
\end{figure}

We collected two datasets of ground mobility robot trajectories: one in a real-world room with everyday objects (e.g., desks, chairs), without any markers, and another in a simulated environment with different obstacles (one at a time), target object, and similar everyday objects. Figures \ref{fig:turtlebot} and \ref{fig:gazebo} show the environments. A trajectory is a sequence of video frames obtained during the robot execution, and each frame is annotated with the expert action taken in the state (translational ($x,y$) and rotational ($\theta$) coordinates). For each trajectory of the lab dataset, a human expert controlled the robot to move toward a target object selected from a total of 7 different objects. For the simulated dataset, the expert controlled the robot to move toward a single target while avoiding an obstacle object selected from 2 different objects. The initial robot placement and the target location were varied leading to varied background scenery. This is a very challenging setting in the aspect that we never explicitly annotate any target or obstacle objects, and there are other distinct everyday objects in the environment such as chairs and other lab equipment.

\subsection{Ablations and Baseline Details}

\begin{figure}[H]
  \centering
    \includegraphics[width=0.48\textwidth]{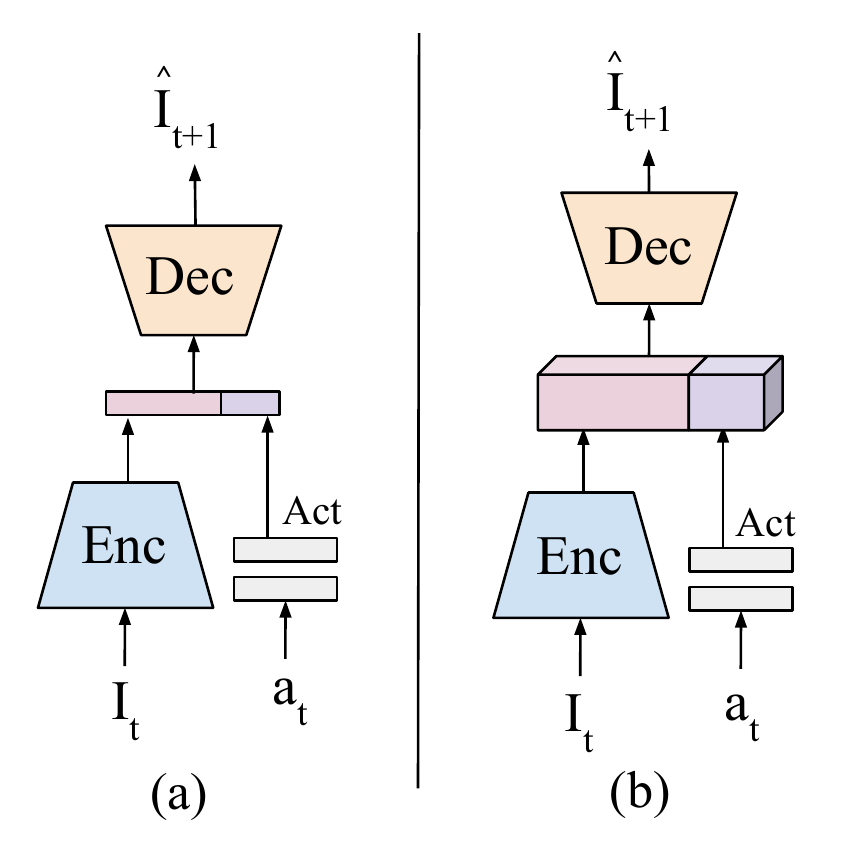}
      \caption{Illustration of the future prediction with a deterministic image predictor. The networks take an image and action as input, concatenate the learned representations, then generate the future image. (a) Linear state representation. (b) Convolutional state representation.}
      \label{fig:models-det}
\end{figure}

To evaluate our design decisions, we compare our full model against other models that replace or remove some of the components. Specifically, we compared using linear state representations against convolutional state representations. Secondly, we examined the impact of the learned prior on our action policy. Without the learned prior, we simply have a deterministic image predictor model that is trained without the latent variable $z_t$. The models resulting from these ablations are shown in Fig.~\ref{fig:models-det}.

In addition to different versions of our approach as mentioned above, we implemented a number of baselines: behavioral cloning, handcrafted critic learning, and forward consistency \cite{pathak2018zeroshot}. Similar to ours, all of these models do not perform any learning from interacting with the environment, as opposed to RL models. Using a CNN formulation, each baseline is described below.

\subsubsection{Behavioral Cloning}
Behavioral cloning (BC) is a straightforward approach to imitation learning, which we formulate as a CNN. The network directly learns a function, $\hat{a}_t = \pi(I_t)$, given current state (i.e., image) as an input and outputs the corresponding action. The BC model is trained using expert samples consisting of a sequence of image-action pairs in a supervised fashion. However, when a given image is not part of the seen expert samples, BC often does not generalize well and produces essentially random actions in those states \cite{laskey2017dart}. This is particularly challenging for real-world robots, where training data is greatly limited. The model is trained to minimize the error between the expert action, $a_t$, and the CNN predicted action, $\hat{a}_t$, using the $L_1$ distance:
\begin{equation} 
 \mathcal{L_B}(a_t, \hat{a}_t) = |a_t - \hat{a}_t|
\end{equation}
During inference, the current image is used as an input to the network $\pi$ and the predicted action is taken by the robot.

\subsubsection{Handcrafted Critic}
Borrowing the idea of taking advantage of non-expert actions behind our future similarity model, we trained a handcrafted critic CNN based on expert angle change as another baseline to distinguish state-action pairs that look like expert samples from non-expert samples. This critic allows the selection of the optimal action given a state by enumerating through action candidates, just like our main approach. This model is simpler in that we formulate this as supervised learning and we also handcraft the `shape' of the outputs.

Given a current image $I_t$ (serving as the state), we apply a CNN $C$ to obtain its vector representation. In parallel, we use a fully-connected neural network, $Act$, to produce a vector representation of the action $a_t$. The two vectors are concatenated and several fully-connected layers follow to produce the critic for state-action pairs:
\begin{equation}
   \hat{Q}(I_t,a_t) = F([C(I_t), Act(a_t)])
\end{equation}

We train the critic function $\hat{Q}$ in a supervised fashion by handcrafting its target values based on the angular distance between the expert action and each candidate action. In this approach, we use the change in angular pose $\delta\theta$ as our action.
Although the translational change $(\delta{x}, \delta{y})$ can be independent of angular change, for simplicity, we set the proposed translational change as a function of angular change $\delta\theta = \arctan(\delta{y} / \delta{x})$ so that there would be only a single parameter to learn. That is, we define our supervision signal to be:
\begin{equation}
\label{eq:reward}
    Q(I_t,a_t) = -|\delta\theta_{exp} - \delta\theta_{a_t}|
\end{equation}
where $\delta\theta_{exp}$ is the expert action and $\delta\theta_{a_t}$ is the candidate action.

Unlike behavioral cloning, this critic function benefits from the training data with non-expert actions while still only taking advantage of offline samples. That is, the training data for the critic CNN may contain pairings of expert state and non-expert actions. This provides the network with significantly more data, allowing more reliable learning even without performing any robot trials. The network is trained to minimize the $L_1$ error between the critic CNN output, $\hat{Q}(I_t,a_t)$ and the (handcrafted) ground truth state-action value, $Q(I_t,a_t)$.
\begin{equation}
\begin{split}
    \mathcal{L_Q} &= |\hat{Q}(I_t,a_t) - Q(I_t,a_t)|\\
     &= |F([C(I_t), Act(a_t)]) - Q(I_t,a_t)|
\end{split}
\end{equation}
During inference, we select the action that gives the maximum critic value (i.e., the action that is expected to be most similar to the expert action):
\begin{equation}
    a_t = \argmax_a \hat{Q}(I_t,a)
\end{equation}

Since our action space is continuous, we randomly sample candidate actions from a uniform distribution from -30 to +30 degrees relative to the robot's current pose to learn the critic function. During inference, we evaluate each of the candidate actions for the current state, and select the one that maximizes the critic function. For a higher dimensional pose change such as by an aerial vehicle, the candidates would be sampled from a spherical cap instead of an arc for a ground mobility robot, and the handcrafted target value would become much more complex. Therefore, it would be desirable to have a critic function that does not require any handcrafting.

\subsubsection{Forward Consistency}

We also experimentally compared our proposed method against the model with Forward Consistency loss from Pathak et al. \cite{pathak2018zeroshot}. Our setting is similar to Pathak et al. \cite{pathak2018zeroshot} in that both learn state-transition models first based on off-policy trajectories and then learn the action policy network (i.e., GSP in \cite{pathak2018zeroshot} and future image similarity critic in ours) without any expert action information by relying only on expert images. The difference is that they used exploration trajectories for the training of their transition model while we used fewer expert trajectories. We took advantage of the code from the authors of the paper, and made it compatible with our problem setting by feeding expert trajectories instead of exploration trajectories for the training of their state transition model. Since this method also requires the goal image of the task as an input, we provided such additional data to its input.

\subsection{Additional Experimental Results and Illustrations}

\begin{figure*}
  \centering
    \includegraphics[width=0.48\textwidth]{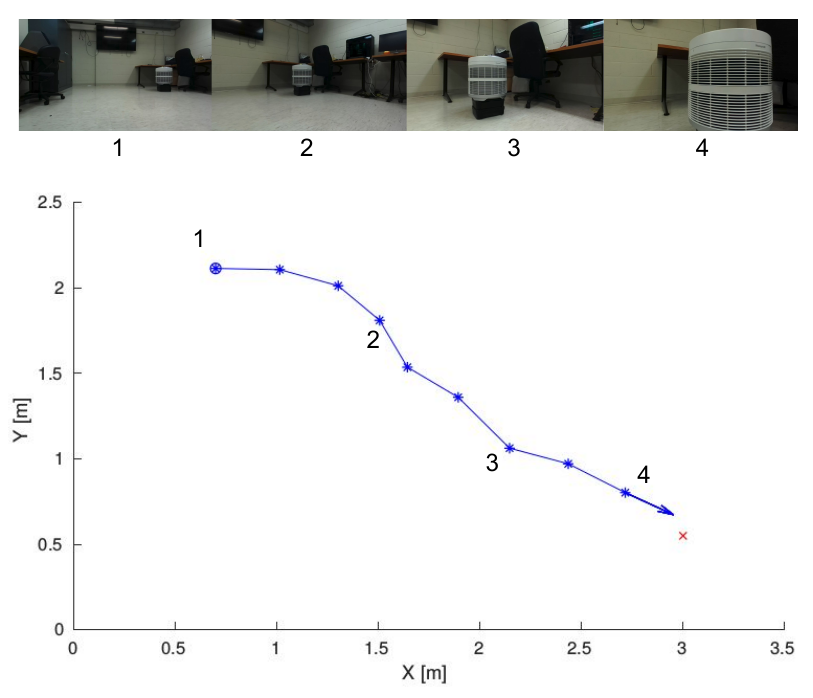}
    \includegraphics[width=0.48\textwidth]{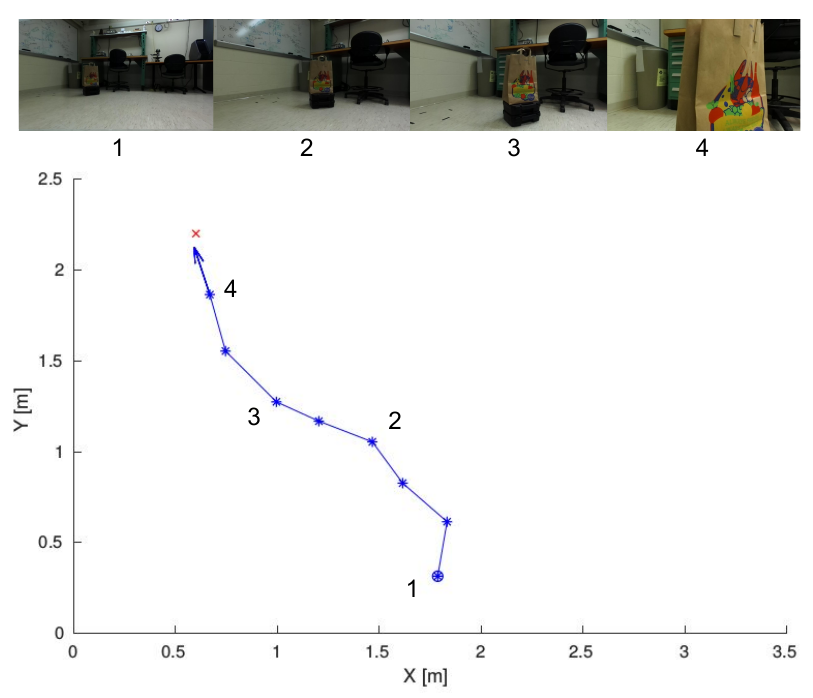}
      \caption{Examples of successful trajectories with the visible frames. The objective of the robot was to approach the air filter (left) and grocery bag (right) target objects.}
      \label{fig:robot-traj-success-1}
\end{figure*}

Fig.~\ref{fig:robot-traj-success-1} illustrates the frames our robot obtains and processes during its task, and Fig.~\ref{fig:robot-traj-example} explicitly compares the robot trajectories of 5 different models. Each model requires less than 30ms to process a single image, measured using a single Titan X (Maxwell) GPU, which allows us run our experiments using the Jetson TX2 mounted on our mobile robot.

Table~\ref{tab:robot-lab2} and Table~\ref{tab:robot-gaz2} report the accuracies of our method against different baselines, including the handcrafted critic. We are able to clearly observe the superiority of our method compared to previous methods.

\begin{table}
\caption{Real-world robot target approach experiment results reporting the task success rate.}
\label{tab:robot-lab2}
\centering
\setlength\tabcolsep{2pt}
\centering
\begin{tabular}{lcccc}
\toprule
Model & Targ1 & Targ2 & Targ3 & Mean \\
\midrule
Behavioral Cloning  & 50\%  & 33\%  & 28\% & 37\% \\
\midrule
Forward Consistency  & 50\%  & 39\%  & 28\%  & 39\% \\
\midrule
Critic Learning-Hand & 67\%  & 61\%  & 50\%  & 59\% \\
\midrule
Det. Linear State    & 67\%  & 56\%  & 56\% & 59\% \\
Det. Conv State      & 78\%  & 78\%  & 67\% & 74\% \\
Stoch. Linear State  & 94\%  & 89\%  & 89\% & 91\% \\
Stoch. Conv State    & 100\% & 89\%  & 94\% & 94\% \\ 
\bottomrule
\end{tabular}
\end{table}

\begin{table}
\caption{Simulated robot obstacle avoidance experiment results reporting the task success rate.}
\label{tab:robot-gaz2}
\centering
\setlength\tabcolsep{2pt}
\begin{tabular}{lccc}
\toprule
Model & Obst.1 & Obst.2 & Mean \\
\midrule
Behavioral Cloning   & 41\%  & 48\%  & 44\% \\
\midrule
Forward Consistency   & 37\%  & 41\%  & 39\% \\
\midrule
Critic Learning-Hand  & 59\%  & 52\%  & 55\% \\
\midrule
Deterministic State  & 44\%  & 37\%  & 41\% \\
Stochastic State     & 67\%  & 59\%  & 63\% \\
\bottomrule
\end{tabular}
\end{table}

\begin{figure*}
  \centering
    \includegraphics[width=0.85\textwidth]{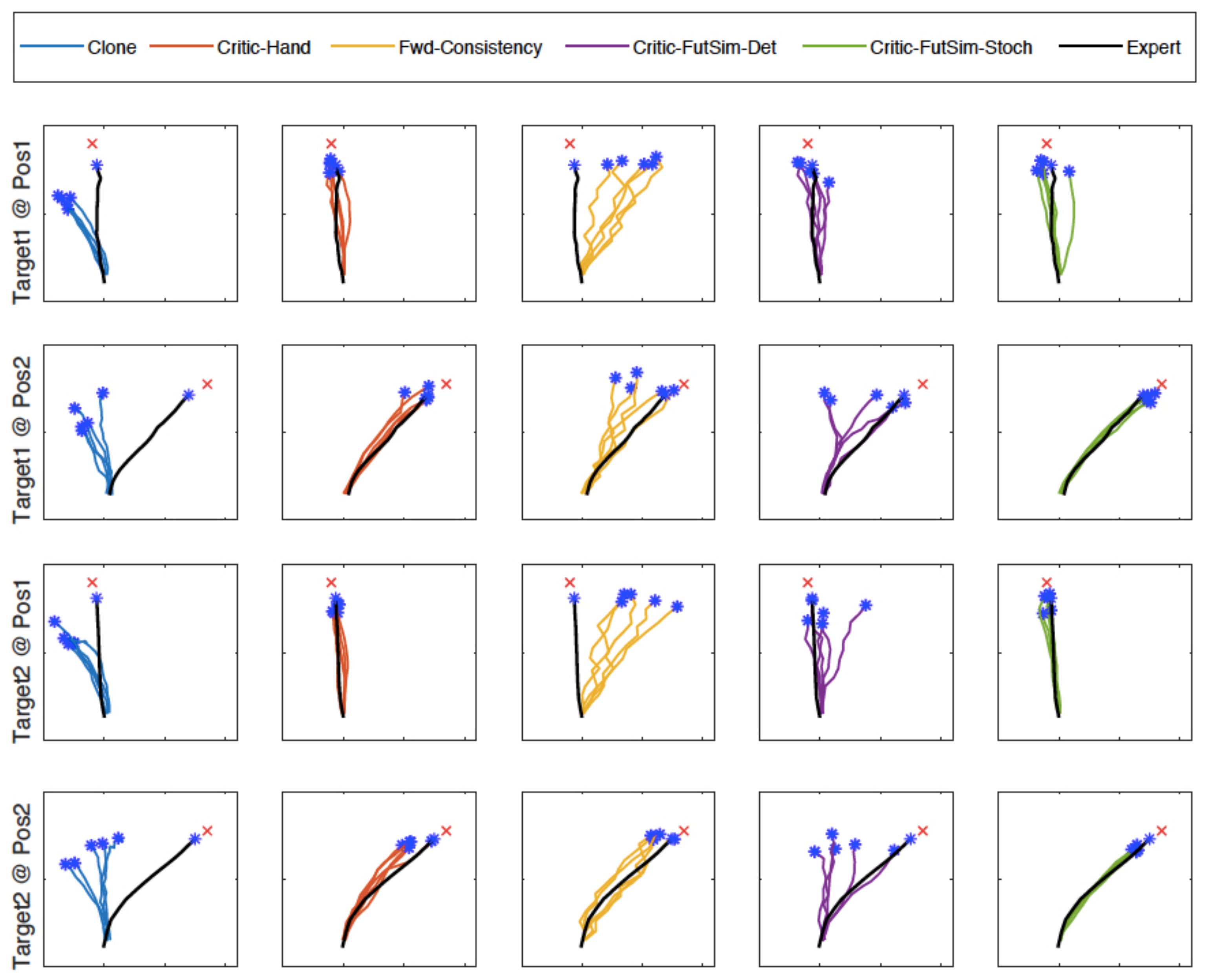}
      \caption{Runs showing the stability of the methods evaluated. Five runs of each method are shown in each box overlaid by the expert trajectory in black. Our method is very stable.}
      \label{fig:stability}
\end{figure*}

To assess the stability of the various methods, we show multiple rollouts of several robot-target configurations in Fig.~\ref{fig:stability}. We ran each method five times for each robot-target placement of two different target objects, including the handcrafted critic. Each configuration is overlaid by the expert trajectory in black. Qualitatively, we see that the stochastic model was the most stable. Table~\ref{tab:similarity2} shows that our method achieved the best similarity score.

\begin{table}
\caption{Similarity scores of various models, comparing their generated trajectories to the (unseen) expert trajectory. The lower the better.}
\label{tab:similarity2}
\centering
\begin{tabular}{lcc}
\toprule
Model & DTW Score\\
\midrule
Behavioral Cloning   & 28.26\\
Forward Consistency  & 18.36\\
Critic Learning-Hand & 9.97\\
FutureSim-Det        & 11.22\\
FutureSim-Stoch      & 5.98\\
\bottomrule
\end{tabular}
\end{table}

\begin{figure}[H]
  \centering
    \includegraphics[width=0.48\textwidth]{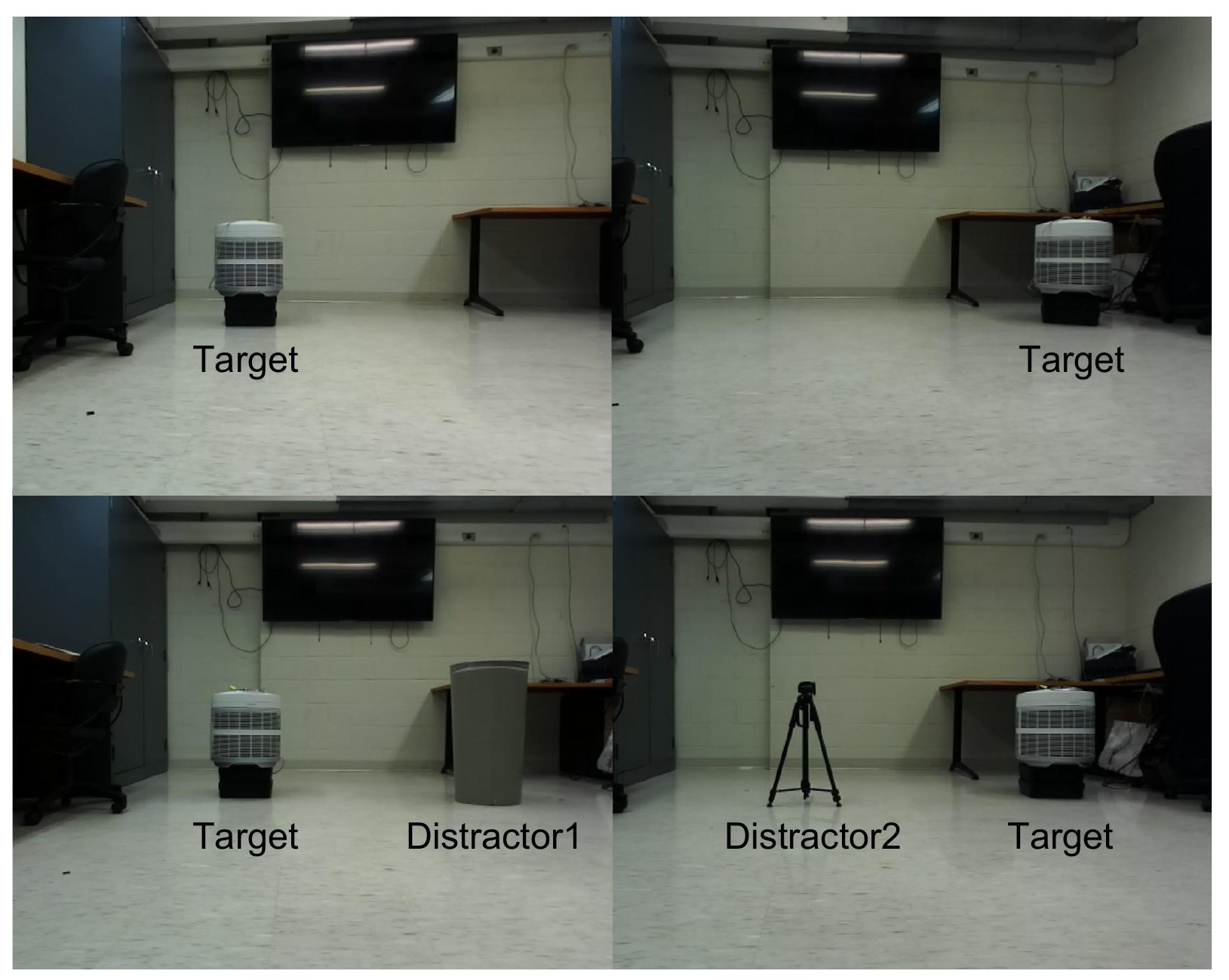}
    \includegraphics[width=0.48\textwidth]{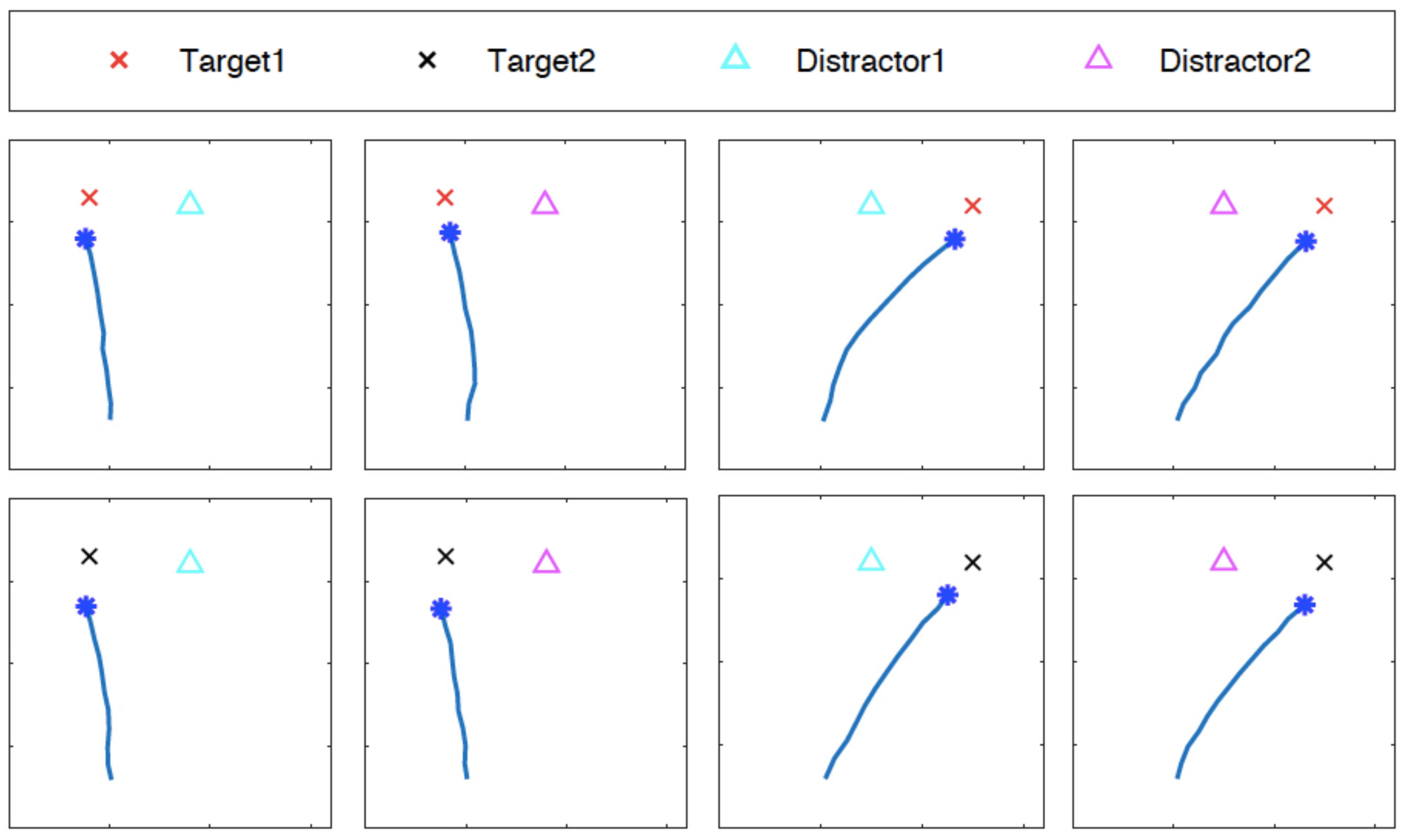}
      \caption{Trajectories with various distractors in place. Our method is immune to distractors, following the same trajectory each time. Left: Example frames without and with the distractor object, showing the distractor is quite large and visually similar to the target object. Right: Trajectories of the distractor test.}
      \label{fig:distractor}
\end{figure}

Our method is robust to distractor objects.  We conducted an experiment where we placed a distractor object in the scene about equidistant to the robot as the target object. We varied the target object, the distractor object, and their locations with respect to the robot. In Fig.~\ref{fig:distractor} (top), we show examples of the target in the lab environment with and without distractors. Even without any labeling of the target or distractor objects, our model is able to ignore the distractor and navigate to the target. A similarity score of 4.94 was achieved for the trajectories shown in Fig.~\ref{fig:distractor} (bottom) using the stochastic model, reflecting consistently small deviation from the expert.

\subsection{Training Information}
Our training curves for the image predictor model and the critic are shown below. For the image predictor of both datasets, we set the learning rate = 0.001 and batch size = 60. The $\beta$ multiplier for the KL loss was set to 0.0001 in our experiments. The learning rate of the value function was set to 5E-6. The weights of the image predictor were held constant when training the value function. We minimize our loss functions with gradient descent using the Adam~\cite{kingma2014adam} solver.

\begin{figure}[H]
    \centering
    \includegraphics[width=0.3\linewidth]{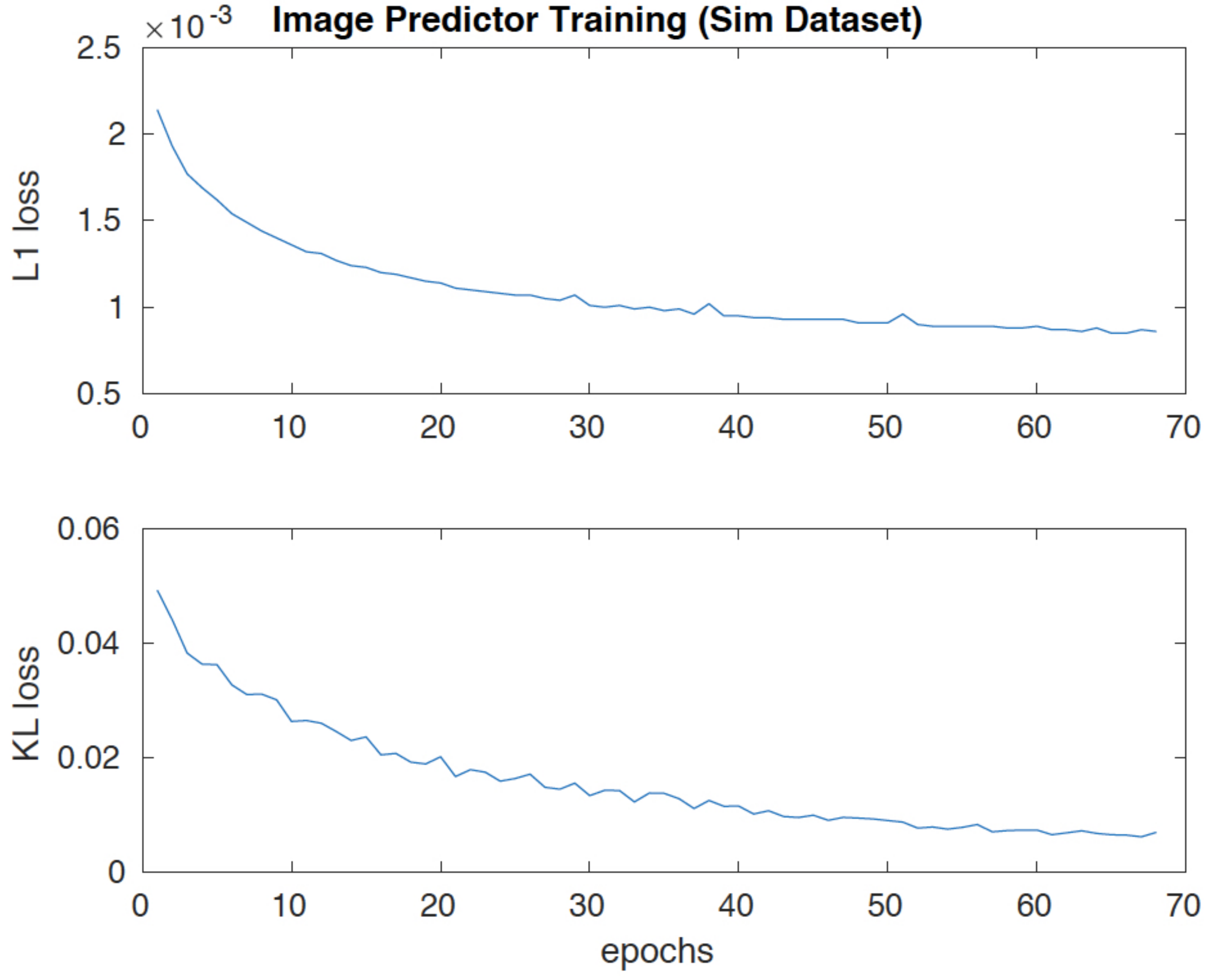}
    \hspace{2mm}
    \includegraphics[width=0.3\linewidth]{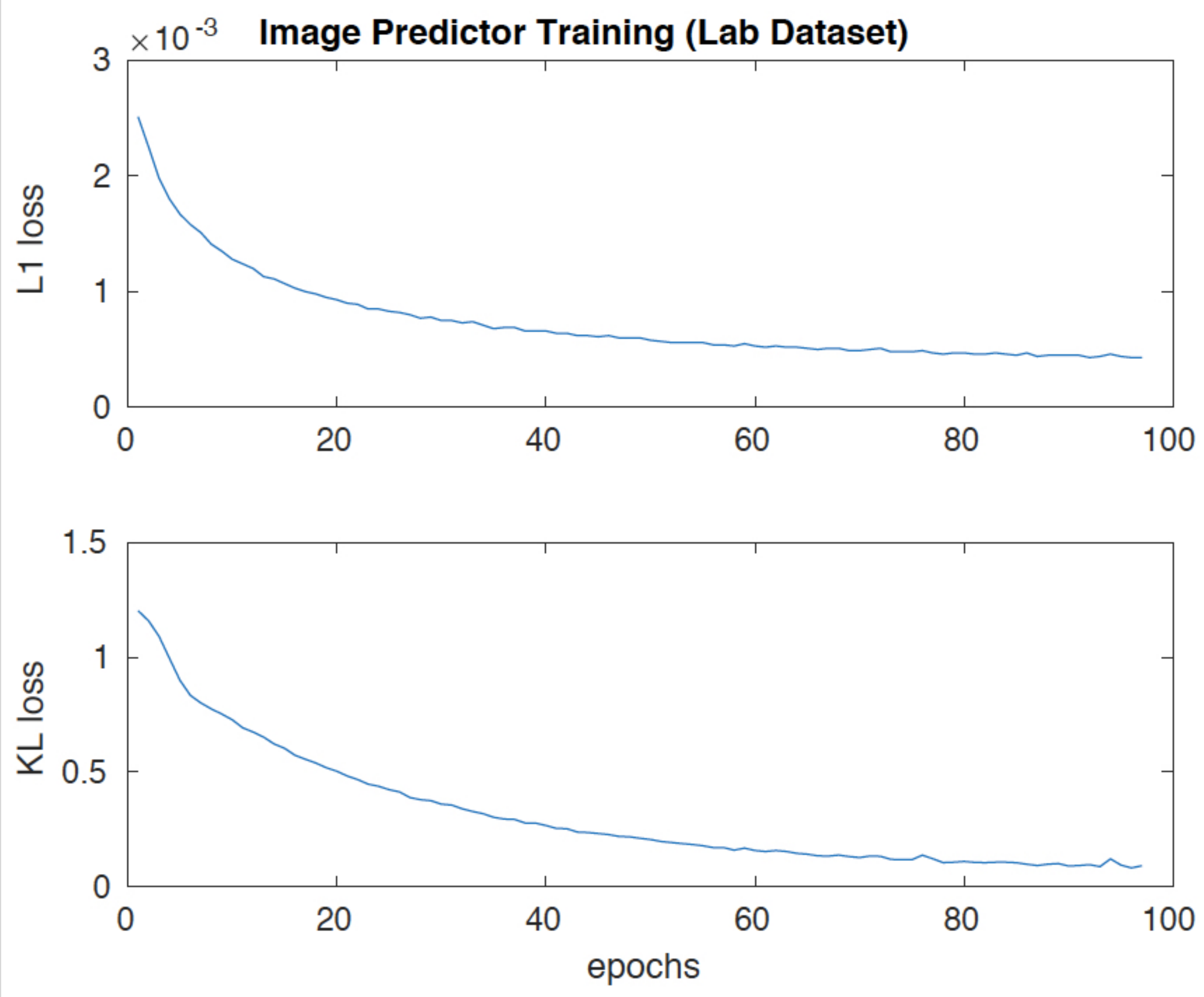}
    \hspace{2mm}
    \includegraphics[width=0.3\linewidth]{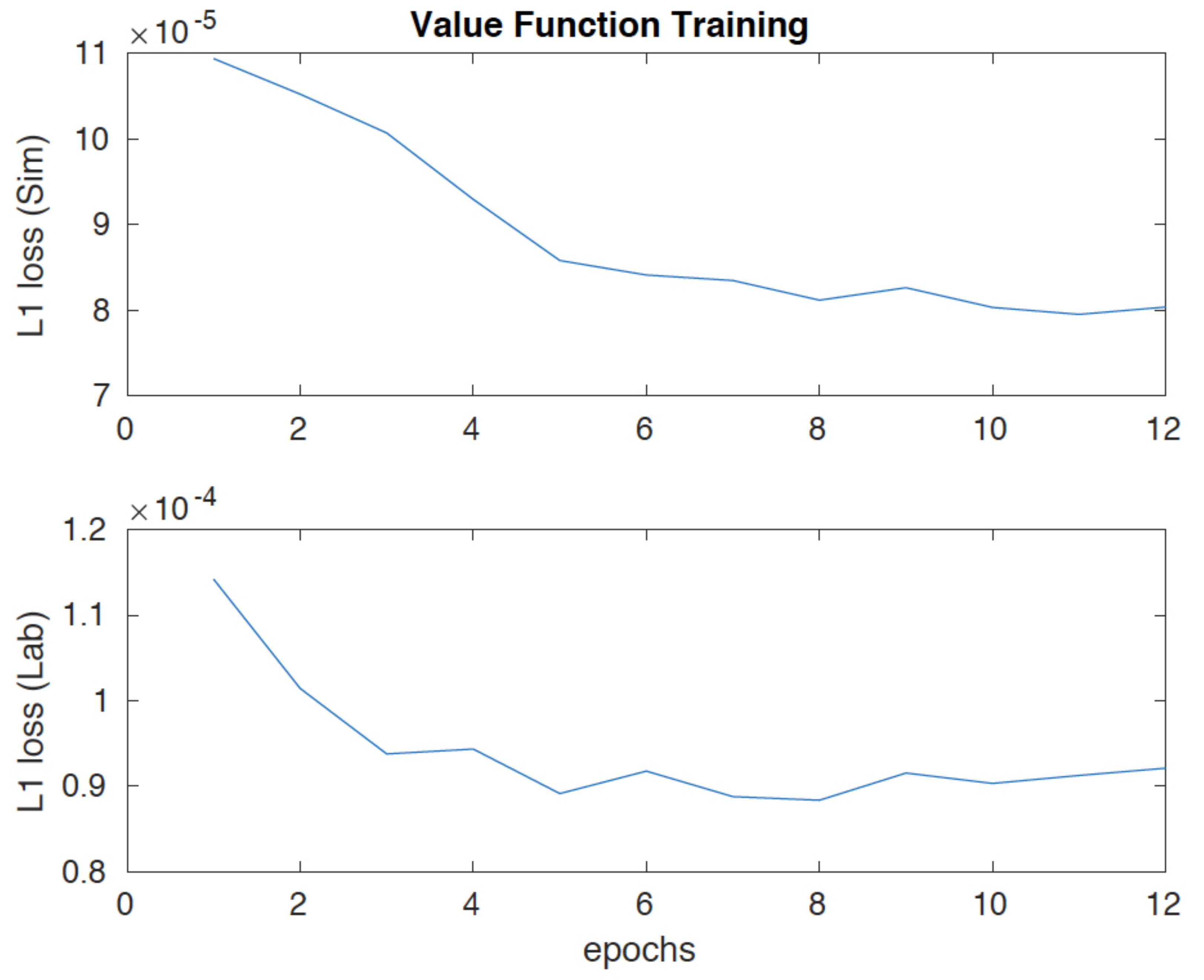}
      \caption{Training curves for the image predictor for the simulated and lab datasets with learning rate = 0.001 and batch size = 60, and the value function with learning rate = 5E-6.}
      \label{fig:training curves}
\end{figure}

\subsection{Dataset and Code}
The project site can be found at https://github.com/anwu21/future-image-similarity

\end{document}